\documentclass[runningheads]{llncs}
\usepackage[T1]{fontenc}
\usepackage{graphicx}

\usepackage{graphicx}

\usepackage{diagbox}
\usepackage[T1]{fontenc}    
\usepackage{booktabs}       
\usepackage{amsfonts}       
\usepackage{nicefrac}       
\usepackage{microtype}      
\usepackage{xcolor}         
\usepackage{times}
\usepackage{epsfig}
\usepackage{graphicx}
\usepackage{amsmath}
\usepackage{amssymb}

\usepackage{amsmath}
\usepackage{amsfonts}
\usepackage{dsfont}
\usepackage{multirow}
\usepackage{adjustbox}
\usepackage{wrapfig}
\usepackage{threeparttable} 

\usepackage{subfigure}
\usepackage{adjustbox} 
\usepackage{tabularx}
\usepackage[colorlinks=true, citecolor=blue, linkcolor=blue, urlcolor=blue]{hyperref}
\usepackage{color} 

\usepackage{pifont}

\begin{document}
\title{Hi-Reco: High-Fidelity Real-Time Conversational~Digital~Humans}
%
%

\author{Hongbin Huang\inst{1} \and
Junwei Li\inst{1} \and
Tianxin Xie\inst{1} \and
Zhuang Li\inst{2} \and
Cekai Weng\inst{1} \and
Yaodong~Yang\inst{1} \and
Yue Luo\inst{1} \and
Li Liu\inst{1} \and
Jing Tang\inst{1} \and
Zhijing Shao\inst{1,2} \and
Zeyu Wang\inst{1,3}}

\authorrunning{Hongbin Huang et al.}
%

\institute{The Hong Kong University of Science and Technology (Guangzhou) \and
Prometheus Vision Technology Co., Ltd. \and
The Hong Kong University of Science and Technology}

\maketitle              
\begin{abstract}
High-fidelity digital humans are increasingly used in interactive applications, yet achieving both visual realism and real-time responsiveness remains a major challenge. We present a high-fidelity, real-time conversational digital human system that seamlessly combines a visually realistic 3D avatar, persona-driven expressive speech synthesis, and knowledge-grounded dialogue generation. To support natural and timely interaction, we introduce an asynchronous execution pipeline that coordinates multi-modal components with minimal latency. The system supports advanced features such as wake word detection, emotionally expressive prosody, and highly accurate, context-aware response generation. It leverages novel retrieval-augmented methods, including history augmentation to maintain conversational flow and intent-based routing for efficient knowledge access. Together, these components form an integrated system that enables responsive and believable digital humans, suitable for immersive applications in communication, education, and entertainment.


\end{abstract}

\section{Introduction}
\label{intro}
Digital humans are increasingly used in communication, education, and entertainment. They play an important role in many downstream applications that involve user interaction, improving user engagement, and delivering personalized experiences. With the rapid development of AI technologies, digital humans are evolving from static avatars into intelligent agents capable of perceiving, understanding, and responding to users in real time.

However, developing a truly high-fidelity digital human that offers a seamless and natural experience remains a significant challenge. A key issue is achieving high realism while preserving natural engagement. Appearance, voice, responses, and motion must feel convincingly human without triggering the uncanny valley effect~\cite{uncannyvalley}. This requires precise control over facial expressions, speech prosody, and non-verbal gestures to enhance believability.

Another major challenge is minimizing system latency to support real-time interaction. Digital humans must respond promptly to users to maintain the flow of conversation. However, delays can arise from processing tasks such as speech synthesis, response generation, and animation, which may operate on different timelines. These delays can interrupt the conversational rhythm and reduce user immersion.

This paper presents \textbf{Hi-Reco}, a high-fidelity real-time conversational digital human system. To address the challenges of realism and responsiveness, our system integrates photorealistic 3D modeling, persona-driven speech synthesis, intelligent response generation, and an asynchronous execution pipeline. By combining lifelike visuals, expressive speech, and context-aware dialogue, the system delivers an engaging experience while mitigating the uncanny valley effect, ensuring real-time performance, and supporting domain-specific knowledge retrieval. In summary, our main contributions are as follows:

\begin{itemize}
    \item \textbf{High-Fidelity Digital Human Interaction.} We develop a photorealistic digital human by integrating advanced 3D modeling, lifelike facial animations, and expressive gestures. Our persona-driven text-to-speech (TTS) system enhances naturalness by generating expressive speech that seamlessly synchronizes with facial movements.
    
    \item \textbf{Knowledge-Grounded Conversational Intelligence.} We introduce a modular retrieval-augmented generation (RAG) framework optimized for domain-specific, real-time interaction. This framework incorporates intent-aware retrieval and dialogue history augmentation, ensuring highly accurate and contextually relevant responses.
    
    \item \textbf{Low-Latency Real-Time System Optimization.} We design an asynchronous execution pipeline to efficiently coordinate 3D animation, TTS synthesis, and response generation. This optimization reduces latency, enabling real-time interaction without perceptible delays.
    
\end{itemize}

\section{Related Work}
\label{relatedwork}
\subsection{3D Avatars}


\textbf{3D Head Avatars.} 
3D models offer greater realism and control over the motion of head avatars. The introduction of 3DMM~\cite{3DMM} marked a significant milestone by representing 3D faces with a fixed topology mesh and parameterizing facial variations through shape and texture bases. Later refinements, including FLAME~\cite{flame2017}, FaceScape~\cite{yang2020facescapelargescalehighquality}, and FaceVerse~\cite{wang2022faceversefinegraineddetailcontrollable3d}, improved adaptability across different ethnicities and enhanced facial expression modeling. Imitator~\cite{Imitator} predicts 3D facial animations from speech while maintaining person-specific motion styles, without necessitating full model retraining, which enables a more flexible approach to driving 3D head avatars.

\textbf{3D Full-Body Avatars.} 
For full-body avatars, modeling strategies involve defining canonical representations and establishing efficient posing mechanisms. Traditional methods utilize explicit parameterized surfaces such as SMPL~\cite{SMPL}, SMPL-X~\cite{SMPLX}, and FLAME~\cite{flame2017} to align texture and motion. With the rise of neural rendering techniques such as Neural Radiance Fields (NeRF)~\cite{mildenhall2020nerfrepresentingscenesneural} and 3D Gaussian Splatting (3DGS)~\cite{3dgs}, more recent innovations leverage neural-based techniques, with representations evolving from point clouds~\cite{zheng2023pointavatardeformablepointbasedhead} to radiance fields~\cite{mildenhall2020nerfrepresentingscenesneural} and 3D Gaussian splatting~\cite{shao2024splattingavatarrealisticrealtimehuman}. The shift from mesh textures to dynamic neural representations has driven significant quality improvements. In terms of real-time performance, different approaches exhibit varying levels of efficiency. AnimatableGaussians~\cite{li2024animatablegaussians} operate at approximately 10 frames per second (FPS), whereas DEGAS~\cite{shao2025degas} achieves superior real-time rendering at 30 FPS, making it a compelling choice for high-fidelity real-time applications.

\subsection{Text-to-Speech}

TTS aims to convert text into human-like speech~\cite{xie2024towards}, widely used in applications like personal assistants and robotics.
Neural TTS utilizes deep learning to produce high-quality and controllable speech.
NaturalSpeech 2~\cite{shen2024naturalspeech2} used diffusion and vector quantization to support zero-shot voice cloning by separating speaker and content. VoiceBox~\cite{le2024voicebox} applied flow-matching diffusion for robust speech generation under noise. 
These methods balance efficiency, naturalness, and control through disentangled representations.
Integrating large language models (LLMs) into TTS enhances contextual reasoning and prompt-driven control.
PromptTTS~\cite{guo2023prompttts} encoded attributes like emotion and pitch into text embeddings for mel-spectrogram control. InstructTTS~\cite{yang2024instructtts} unified semantic and acoustic spaces via diffusion, enabling user-friendly speech synthesis over style prompts like ``excited tone.''
CosyVoice~\cite{du2024cosyvoice} used supervised semantic tokens to improve content consistency and speaker similarity. These approaches demonstrate the synergy between LLMs and speech modeling for intuitive, high-fidelity synthesis.

\subsection{Retrieval-Augmented Generation}



The canonical RAG architecture decouples retrieval and generation: a retriever module first identifies top-k relevant passages from a large external corpus via vector or hybrid similarity search, then a generator module fuses this context into a final answer~\cite{lewis2020retrieval,gao2023retrieval}. This explicit grounding enables LLMs to surpass closed-book performance, significantly reducing factual errors in question answering~\cite{wang2023survey}. Systematic analyses reveal that RAG not only mitigates hallucination~\cite{li2024enhancing}, but also dramatically improves LLM responses to domain-specific and underrepresented knowledge, as shown in both generic and vertical-domain benchmarks~\cite{wang2024domainrag}. Recent surveys further outline RAG's versatility across modalities, including code-switched language understanding~\cite{hong2025migrate} and spoken language processing~\cite{yang2024srag}.

\section{Methods}
\label{Methods}
\begin{figure}[htb!]
    \centering
    \includegraphics[width=1.0\linewidth]{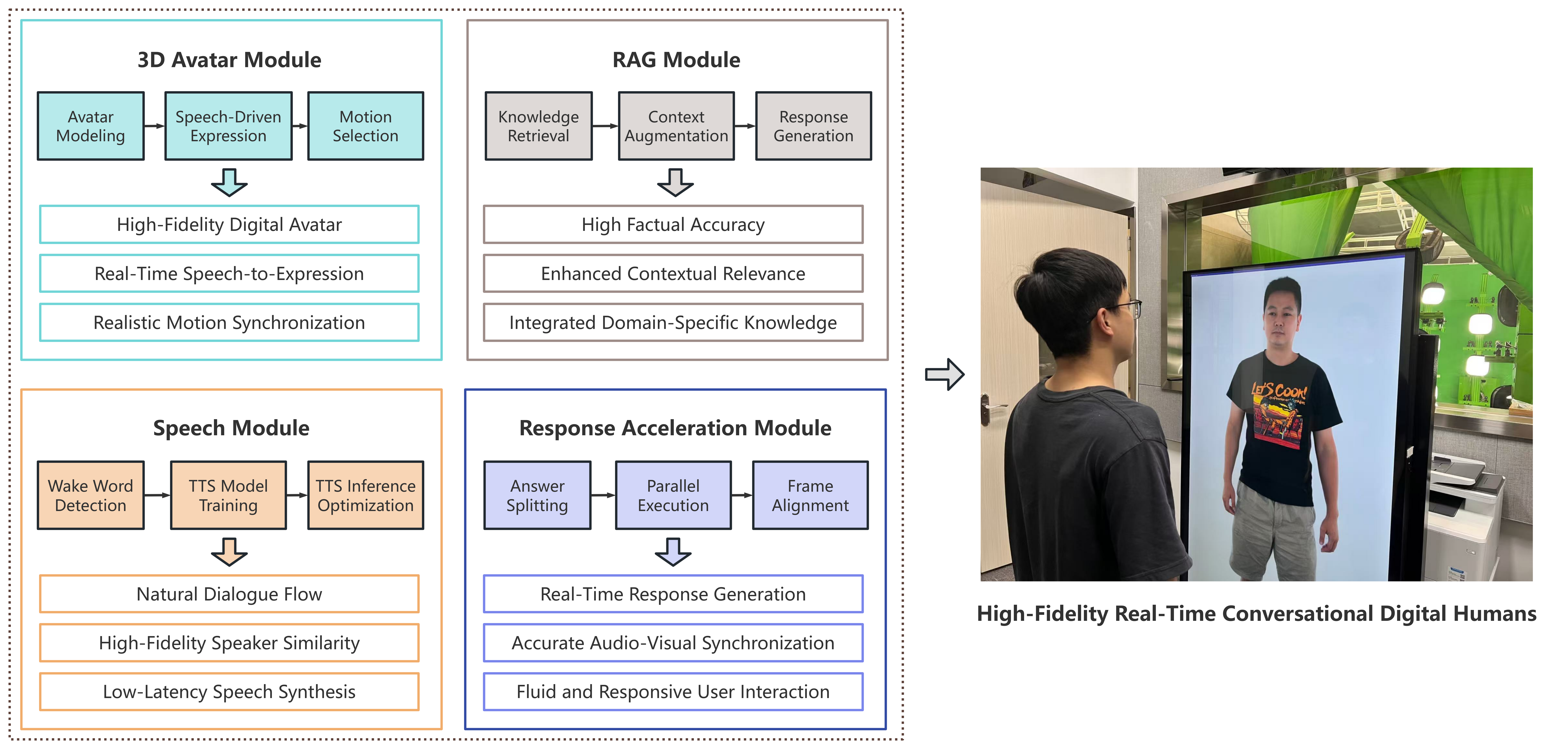}
    \caption{System architecture of our digital human framework.}
    \label{fig:SystemArchitecture}
\end{figure}

Our digital human system consists of several coordinated modules that together enable high-fidelity real-time interaction. This section details the design and functionality of each component, as outlined in Fig.~\ref{fig:SystemArchitecture}. We introduce the 3D avatar module in Section~\ref{sec:avatar_module}, the speech module in Section~\ref{sec:speech_module}, the RAG module in Section~\ref{sec:rag_module}, and the response acceleration module in Section~\ref{sec:latency_opt_module}.

\subsection{3D Avatar Module}
\label{sec:avatar_module}



\textbf{3D Avatar Modeling.} DEGAS~\cite{shao2025degas} demonstrates outstanding performance in both rendering fidelity and real-time frame rate, making it a strong foundation for high-quality avatar animation. In our system, we adopt a 3D avatar modeling framework based on DEGAS to leverage its strengths in photorealistic rendering and coherent body-expression integration. 

However, the original DEGAS pipeline relies on the disentanglement of pose and expression~\cite{pang2023dpedisentanglementposeexpression} to extract facial expression features from synthesized 2D talking face videos. This process involves generating videos from speech and then applying image-based expression extraction, which results in high computational cost and latency. As a result, the pipeline is not suitable for real-time applications. To overcome this limitation, we adopt a more direct and efficient approach by replacing the video-based expression extraction module with a real-time speech-driven solution. Specifically, we use Imitator~\cite{Imitator} to predict FLAME expression parameters~\cite{flame2017} directly from speech input. This design avoids intermediate visual representations, reduces system complexity and latency, and enables temporally coherent and emotionally aligned facial animations. In addition, we accelerate the DEGAS encoder-decoder architecture using TensorRT~\cite{TensorRT}, further improving inference speed and responsiveness in interactive scenarios.



\begin{figure}[htb!]
    \centering
    \includegraphics[width=1.0\linewidth]{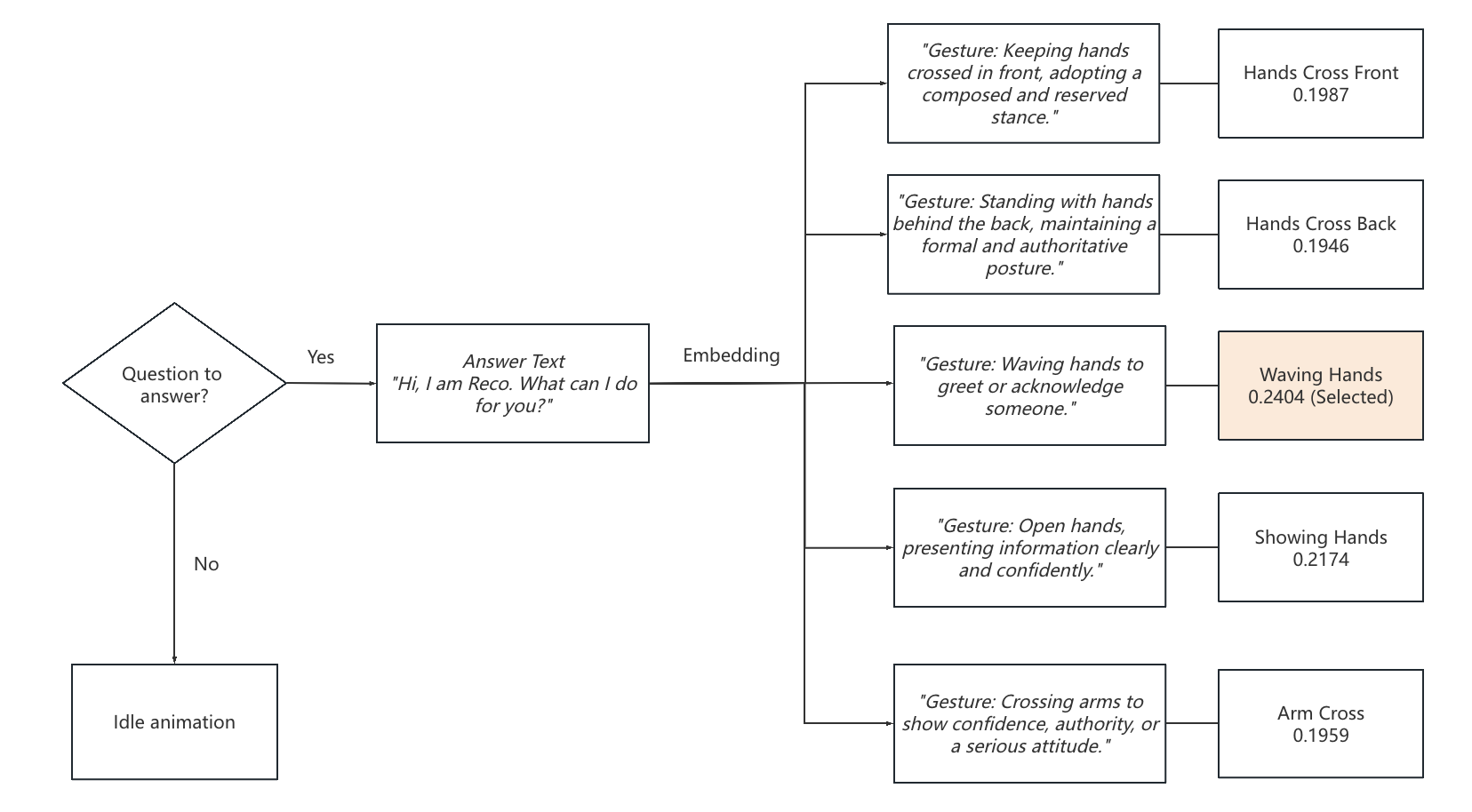}
    \caption{An example of semantic embedding-based motion selection.}
    \label{fig:SelectMotion}
\end{figure}

\textbf{Motion Selection.}
We leverage a pretrained language model, Sentence-BERT~\cite{reimers-2019-sentence-bert}. Each predefined motion is associated with an embedding that captures its typical conversational usage, derived from a manually curated description of the intent of the gesture. Fig.~\ref{fig:SelectMotion} shows an example of motion selection. When generating a response, the system extracts a text embedding from the spoken content and computes the cosine similarity between the response embedding and the motion embeddings. The motion with the highest similarity score is selected and scheduled for playback.

\subsection{Speech Module}
\label{sec:speech_module}




\textbf{Wake Word Detection.} To ensure robust and intentional user interaction in noisy or multi-speaker environments, we leverage Porcupine~\cite{wakeword}, a lightweight and efficient on-device wake word detection engine, to continuously monitor for a predefined keyword (e.g., ``Hi Reco''). Only when the wake word is detected does the system proceed to the speech recognition and response pipeline. 







\textbf{Text-to-Speech.}
We first fine-tune GPT-SoVITS~\cite{gptsovits} using an internally collected dataset and deploy it for real-time streaming inference. Our implementation of GPT-SoVITS focuses on streamlined voice cloning through a structured data preprocessing pipeline.
For vocal isolation, we utilize UVR5~\cite{uvr5} to separate vocals from background noise, ensuring high-quality input. Audio segmentation is automated using silence detection (-40dB threshold) to split recordings into 5-second clips, minimizing contextual interference during training.
Text alignment employs Whisper Large-V3~\cite{radford2023whisper} for multilingual support, with manual verification through an interactive WebUI to correct punctuation and remove noisy segments.
The V2 architecture of GPT-SoVITS further enhances phoneme alignment through a dual-token mechanism (global and semantic tokens), resolving mispronunciation issues observed in earlier versions.
Training requires only 1 minute of clean audio, optimized via joint fine-tuning of SoVITS (variational autoencoder) and GPT (semantic transformer) modules, completing within 10 minutes on consumer-grade GPUs.

Our deployment framework achieves 0.7-second latency through three key innovations.
First, chunk-based processing splits inputs into 150ms segments with overlap-add reconstruction, enabling parallel synthesis while maintaining prosodic continuity.
Second, FP16 quantization reduces SoVITS model size by 40\% without compromising quality (MOS$\geq$4.3 in zero-shot scenarios).
Third, a hybrid API architecture combines RESTful endpoints for batch processing and WebSocket streaming for digital avatar integration, synchronizing synthesized speech with our digital humans.
For edge deployment, TensorRT~\cite{TensorRT} optimization enables 30ms/sentence generation on 4GB GPUs through adaptive CPU-GPU workload balancing.
This system currently supports Chinese and English with automatic code-switching detection, outperforming traditional TTS approaches in both efficiency (5× faster training) and emotional expressiveness.
\subsection{RAG Module}
\label{sec:rag_module}

The RAG module is the intelligence component for generating accurate, knowledge-based responses in our real-time digital human system.  Serving as a core intelligent component, it aims to augment LLMs with information from an external, domain-specific knowledge base, providing accurate, knowledge-based responses for university resources, and ensuring factual accuracy and responsiveness critical for interactive applications. It employs a retrieve-then-generate architecture: its knowledge base comprises campus customer service documentation; the retriever uses LightRAG~\cite{guo2024lightrag} for context-aware chunking and vector indexing; and the generator, a fine-tuned Qwen2.5-32B-Instruct~\cite{qwen2.5} on more than 10k domain-specific campus examples to enhance its understanding and grounding ability and synthesize responses from queries, history, and retrieved passages.


To optimize real-time performance and conversational coherence, two key strategies are implemented: \textbf{1. History-Augmented Retrieval:} Dynamically integrates full dialogue history, significantly enhancing multi-turn retrieval relevance and coherence.  \textbf{2. Intent-Based Routing:} Routes queries via an intent classifier to smaller, domain-specific vector indexes, substantially reducing retrieval latency with minimal accuracy impact.
\begin{figure}[htb!]
    \centering
    \includegraphics[width=1.0\linewidth]{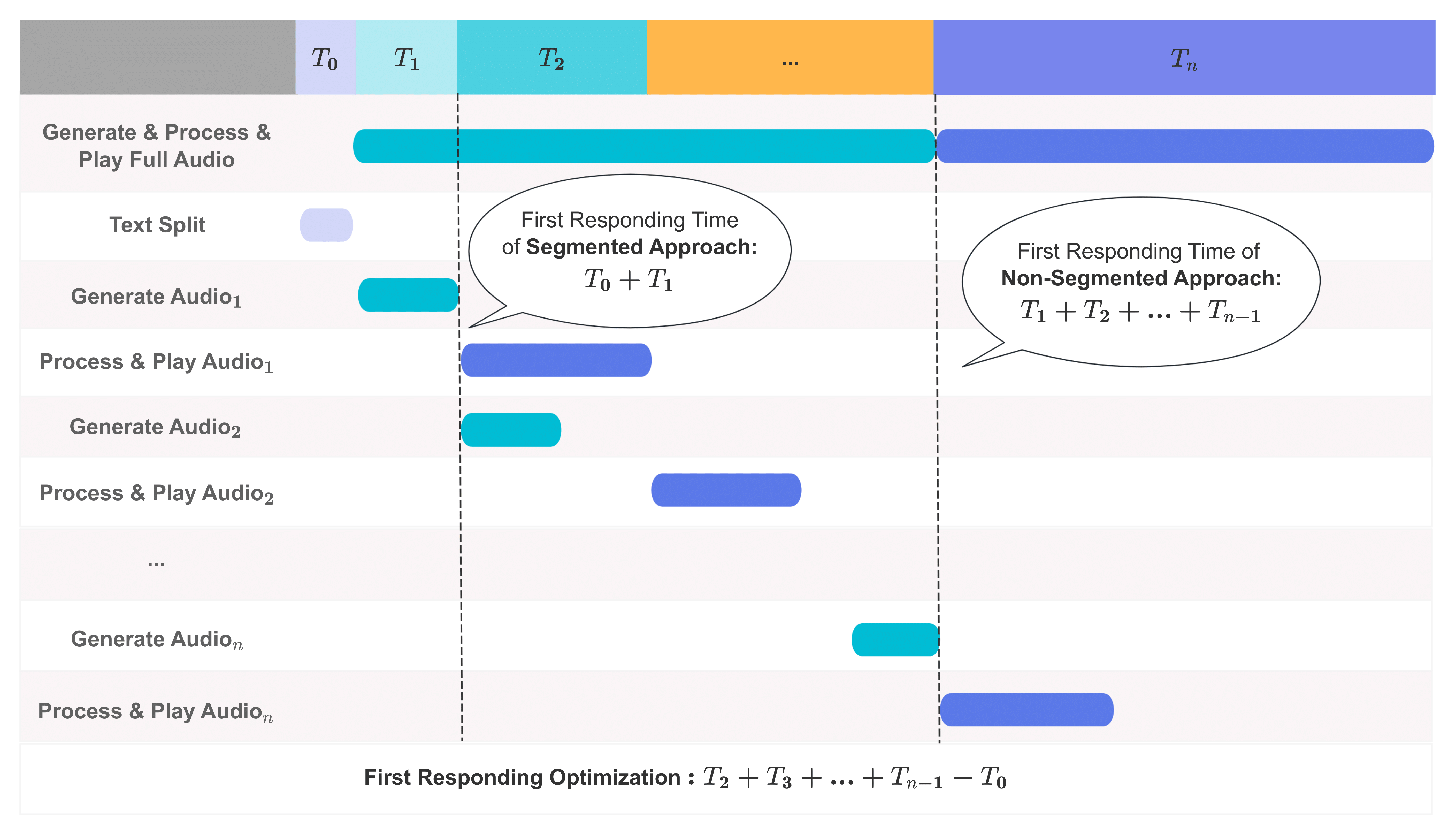}
\caption{Comparison between segmented and non-segmented audio processing pipelines.}
    \label{fig:SystemOpt}
\end{figure}

\subsection{Response Acceleration Module}
\label{sec:latency_opt_module}




\textbf{Latency Optimization.} Fig.~\ref{fig:SystemOpt} shows the comparison between segmented and non-segmented audio processing pipelines. To reduce the latency, we employ the asynchronous execution pipeline. In the case of the TTS module, the system can be optimized by splitting the generated answer into smaller chunks rather than generating the entire response in one go. We split the answer into several segments through the text split process. This split allows the system to immediately begin generating the next segment of audio while processing the first. 






\textbf{Real-Time Synchronization.}
When a new audio segment is generated and played, the module triggers the simultaneous generation of corresponding facial expressions and body motion data. As the system processes audio in chunks, face and motion data are calculated and rendered dynamically to align with the current audio timestamp. The critical factor here is determining which frame of facial and motion data corresponds to the current audio playback position. The index of the visual frame is determined by the audio timestamp (in milliseconds) and FPS, as shown in Equation~\eqref{eq:motionpos}:

\begin{equation}
\mathrm{frame\_index} = \mathrm{audio\_timestamp} \times \mathrm{\frac{\mathrm{fps}}{1000}}
\label{eq:motionpos}
\end{equation}


In summary, the response acceleration module not only minimizes latency but also ensures that the system maintains high fidelity, providing a natural, immersive experience for users interacting with the digital human.

\section{Experiments}
\label{Experiments}
\subsection{3D Avatars}

\begin{figure}[htb!]
  \centering
  \includegraphics[width=1.0\linewidth]{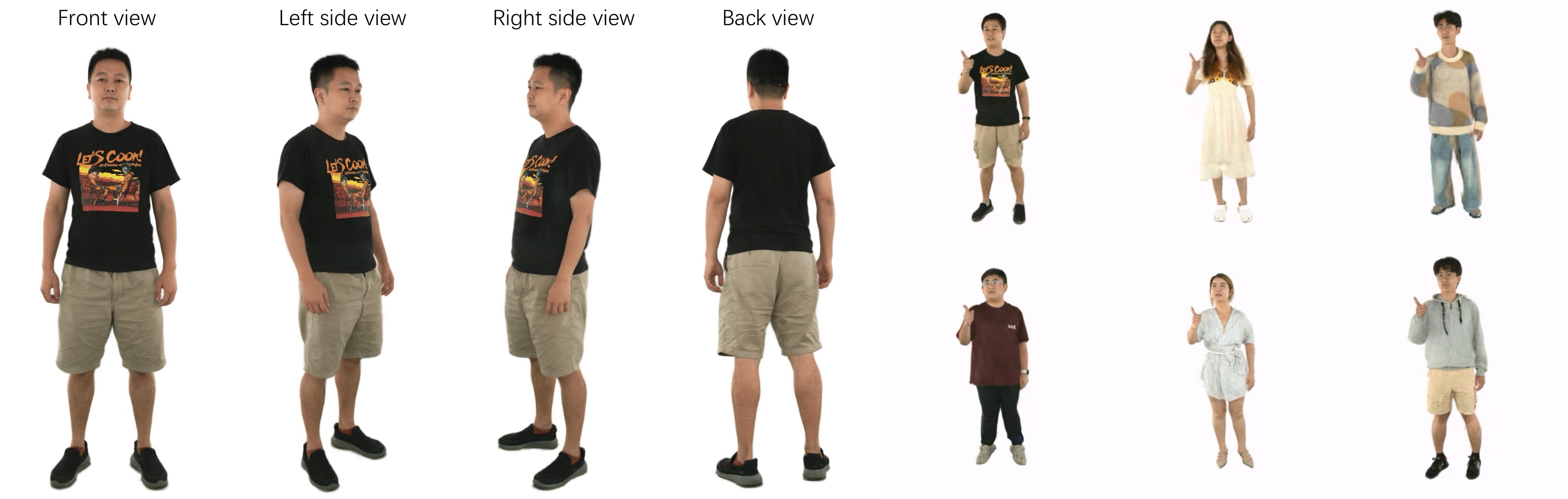}
  \caption{Rendering results of the 3D digital human avatar from four canonical views and diverse 3D digital humans.}
  \label{fig:RenderResult}
\end{figure}

While the 3D digital human model allows for full 360-degree inspection, we generated diverse digital humans with different appearances by showcasing four canonical views: front, left side, right side, and back (Fig.~\ref{fig:RenderResult}). To further enhance the system's expressiveness and real-time capability, we adopt a speech-driven expression modeling strategy~\cite{Imitator} that directly generates FLAME expression parameters~\cite{flame2017} from audio. Our method eliminates the need for intermediate video synthesis, enabling efficient and semantically aligned facial animation from speech input.


\subsection{Text-to-Speech}

We evaluated our TTS model using two key metrics: Word Error Rate (WER) for intelligibility and Similarity Mean Opinion Score (S-MOS) for perceived quality and speaker similarity. WER reflects ASR transcription accuracy, while S-MOS captures human judgments of naturalness and resemblance to reference speech. We compared our model against CosyVoice~\cite{du2024cosyvoice} and F5-TTS~\cite{chen2024f5}, using both objective and subjective evaluations. Each system generated 20 samples (10 Mandarin, 10 English) from phonetically and linguistically diverse sentences.

\textbf{Objective Evaluation.}
We used Whisper (Large)~\cite{radford2023whisper} to transcribe synthesized speech and computed the average WER over all samples. Table~\ref{tab:tts_eval} shows WER results for Mandarin-to-Mandarin (zh→zh) and Mandarin-to-English (zh→en) voice cloning. F5-TTS achieved the lowest WER in both directions (0.074, 0.018). CosyVoice performed better than our model in zh→zh (0.078 vs. 0.112) but worse in zh→en (0.139 vs. 0.027). Our method shows strong cross-lingual performance but lags in the monolingual case.
However, WER alone cannot capture aspects like prosody or speaking style. Models with similar WERs may differ significantly in expressiveness—e.g., flat speech can yield low WER but poor naturalness. Therefore, subjective evaluation is essential.

\textbf{Subjective Evaluation.}
For the subjective evaluation, we invited five human raters to evaluate 20 samples per system, scoring naturalness, intelligibility, and speaker similarity on a 1–5 scale. Final S-MOS scores were averaged across raters and samples for each direction.
As shown in Table~\ref{tab:tts_eval}, our model achieved the highest S-MOS in both zh→zh (4.310) and zh→en (4.280), outperforming F5-TTS (4.060 in zh→zh) and CosyVoice (lowest in both). These results confirm that our method delivers superior perceived quality and cross-lingual speaker consistency.


\begin{table}[t]
\caption{Objective (WER) and subjective (S-MOS) comparison under two conditions (zh$\rightarrow$zh and zh$\rightarrow$en) between our TTS method and existing SOTA methods.}
\label{tab:tts_eval}
\centering
\renewcommand{\arraystretch}{1.2}
\begin{tabular}{|l|l|l|l|l|}
\hline
\multirow{2}{*}{Method} & \multicolumn{2}{l|}{WER $\downarrow$} & \multicolumn{2}{l|}{S-MOS $\uparrow$} \\
\cline{2-5}
                        & zh$\rightarrow$zh & zh$\rightarrow$en & zh$\rightarrow$zh & zh$\rightarrow$en \\
\hline
CosyVoice~\cite{du2024cosyvoice} & 0.078 & 0.139 & 3.340 & 3.030 \\
F5-TTS~\cite{chen2024f5}         & \textbf{0.074} & \textbf{0.018} & 4.060 & 3.440 \\
\hline
Ours                             & 0.112 & 0.027 & \textbf{4.310} & \textbf{4.280} \\
\hline
\end{tabular}
\end{table}

\subsection{Retrieval-Augmented Generation}



\textbf{History‐Augmented RAG Retrieval.}
This experiment aimed to evaluate the improvement in retrieval relevance for follow-up queries by augmenting the retrieval corpus with full multi-turn chat history. We compared two RAG variants: Static RAG, which uses only the original campus customer service corpus, and Dynamic RAG, which augments that corpus with concatenated dialogue history. A BGE-reranker-based model was utilized to measure the Top-1 score on 10 representative follow-up queries. As shown in Table~\ref{tab:speech_to_expression_comparison}, the Dynamic RAG variant improved the average Top-1 score by 43.3\% over Static RAG, demonstrating that including complete multi-turn conversational context as supplementary knowledge enhances retrieval accuracy and relevance.  

\begin{table}[ht]
\caption{Top-1 reranker scores of static and dynamic RAG.}
\label{tab:speech_to_expression_comparison}
\centering
\renewcommand{\arraystretch}{1.2}
\begin{tabular}{|l|l|}
\hline
Method & Top-1 Score $\uparrow$ \\
\hline
Static RAG & 5.1916 \\
Dynamic RAG & \textbf{7.4388} \\
\hline
\end{tabular}
\end{table}




\textbf{Intent-Based Routing and Latency Reduction.}
Regarding Intent-Based Routing and Latency Reduction, an experiment investigated improving retrieval efficiency in domain-specific vector indices while maintaining response quality. Three configurations were compared: a unified index, a random routing strategy, and an intent routing method using a classifier to direct queries to relevant indices. The test set comprised 500 queries from five domains (Academics, Finance, Campus Services, IT, and Library), with latency and Top-1 score reported. As shown in Table~\ref{tab:retrieval-comparison},  Category-based retrieval (i.e., intent-based routing) achieved a 35.2\% reduction in latency with minimal impact on Top-1 accuracy, proving to be an effective strategy for enhancing both speed and usability in modular QA systems.

\begin{table}[h]
\caption{Comparison of latency and score between full corpus retrieval and category-based retrieval.}
\label{tab:retrieval-comparison}
\centering
\footnotesize
\renewcommand{\arraystretch}{1.2}
\begin{tabular}{|l|l|l|}
\hline
Method & Latency [s] $\downarrow$ & Top-1 Score $\uparrow$ \\
\hline
Full Corpus Retrieval & 6.019 & \textbf{4.8576} \\
Category-Based Retrieval & \textbf{3.902} & 4.0100 \\
\hline
\end{tabular}
\end{table}
\vspace{-1em}



\subsection{Response Acceleration}





We conducted an experiment by comparing the time to first audio playback under two different scenarios, using the same response to control for variables. The two approaches are described as follows:

\begin{enumerate}
    \item \textbf{Non-Segmented Approach:} The entire response was passed to the TTS system at once.
    \item \textbf{Segmented Approach:} The response was split into smaller phrases, where each phrase was fed to the TTS system individually. As soon as the first phrase's audio was generated, playback began while the remaining phrases were still being processed.
\end{enumerate}

The results of the experiment are summarized in Table~\ref{tab:latency_results}.
From the table, we observed that the use of the segmented approach significantly reduces the time to the first audio playback, resulting in an 85.0\% reduction in initial latency. This confirms that splitting the sentence into smaller phrases and processing them in parallel can effectively improve user experience by providing a faster initial response while maintaining natural continuity in speech synthesis.

\begin{table}[ht]
\caption{Comparison of time to first audio playback between segmented and non-segmented TTS processing.}
\label{tab:latency_results}
\centering
\renewcommand{\arraystretch}{1.2}
\begin{tabular}{|l|l|}
\hline
Method & Time to First Audio Playback [s] $\downarrow$ \\
\hline
Non-Segmented & 10.223 \\
Segmented & {\bfseries 1.538} \\
\hline
\end{tabular}
\end{table}

\subsection{Limitations and Future Work}

Our system relies on a fixed library of gestures selected by dialogue context, rather than generating motion in real time from speech. While effective for basic expressiveness, this approach limits the digital human's ability to produce body language that accurately reflects the spoken language's rhythm, dynamics, and emotional tone. As a result, gestures may sometimes appear repetitive, mismatched, or less responsive to subtle variations in speech delivery. 

Future systems could address these limitations by generating motion sequences directly from speech audio, incorporating features such as prosody, intonation, and emotional cues. This would enable more adaptive, expressive, and context-aware non-verbal behaviors, leading to digital humans that can engage users with greater naturalness and emotional authenticity.

\section{Conclusion}
\label{Conclusion}
This paper has presented a high-fidelity, real-time conversational digital human system that seamlessly integrates visual realism with intelligent responsiveness. We demonstrated that a tightly coordinated pipeline spanning 3D avatar modeling, speech processing, and knowledge-grounded dialogue forms the foundation for believable and engaging virtual agents. Our approach supports advanced features, including persona-driven speech synthesis for enhanced emotional expressiveness and a modular RAG framework, which leverages intent-aware retrieval and dialogue history augmentation to improve accuracy and contextual relevance. We further showed how system-level optimizations, including latency reduction and real-time synchronization, significantly enhance the smoothness of user interaction. Together, these components push the boundaries of interactive digital humans, offering new opportunities for immersive applications in communication, education, and entertainment.

%
%
%
%

\bibliographystyle{splncs04}
\bibliography{main}

\begin{thebibliography}{10}
\providecommand{\url}[1]{\texttt{#1}}
\providecommand{\urlprefix}{URL }
\providecommand{\doi}[1]{https://doi.org/#1}

\bibitem{uvr5}
Anjok07: {Ultimate Vocal Remover GUI v5.6}. \url{https://github.com/Anjok07/ultimatevocalremovergui} (2025), accessed: 2025-03-20

\bibitem{3DMM}
Blanz, V., Vetter, T.: {A Morphable Model for the Synthesis of 3D Faces}. In: Proceedings of the 26th Annual Conference on Computer Graphics and Interactive Techniques. p. 187–194. SIGGRAPH '99, ACM Press/Addison-Wesley Publishing Co., USA (1999). \doi{10.1145/311535.311556}

\bibitem{chen2024f5}
Chen, Y., Niu, Z., Ma, Z., Deng, K., Wang, C., Zhao, J., Yu, K., Chen, X.: {F5-TTS: A Fairytaler that Fakes Fluent and Faithful Speech with Flow Matching}. arXiv preprint \href{https://arxiv.org/abs/2410.06885}{arXiv:2410.06885}  (2024)

\bibitem{du2024cosyvoice}
Du, Z., Chen, Q., Zhang, S., Hu, K., Lu, H., Yang, Y., Hu, H., Zheng, S., Gu, Y., Ma, Z., et~al.: {CosyVoice: A Scalable Multilingual Zero-shot Text-to-speech Synthesizer based on Supervised Semantic Tokens}. arXiv preprint \href{https://arxiv.org/abs/2407.05407}{arXiv:2407.05407}  (2024)

\bibitem{gao2023retrieval}
Gao, Y., Xiong, Y., Gao, X., Jia, K., Pan, J., Bi, Y., Dai, Y., Sun, J., Wang, H., Wang, H.: {Retrieval-Augmented Generation for Large Language Models: A Survey}. arXiv preprint \href{https://arxiv.org/abs/2312.10997}{arXiv:2312.10997}  (2023)

\bibitem{guo2023prompttts}
Guo, Z., Leng, Y., Wu, Y., Zhao, S., Tan, X.: {PromptTTS: Controllable Text-to-Speech with Text Descriptions}. In: IEEE International Conference on Acoustics, Speech and Signal Processing. pp.~1--5 (2023)

\bibitem{guo2024lightrag}
Guo, Z., Xia, L., Yu, Y., Ao, T., Huang, C.: {LightRAG: Simple and Fast Retrieval-Augmented Generation}. arXiv preprint \href{https://arxiv.org/abs/2410.05779}{arXiv:2410.05779}  (2024)

\bibitem{hong2025migrate}
Hong, S., Lee, S., Moon, H., Lim, H.S.: {MIGRATE: Cross-Lingual Adaptation of Domain-Specific LLMs through Code-Switching and Embedding Transfer}. In: Proceedings of the 31st International Conference on Computational Linguistics. pp. 9184--9193 (2025)

\bibitem{3dgs}
Kerbl, B., Kopanas, G., Leimk{\"u}hler, T., Drettakis, G.: {3D Gaussian Splatting for Real-Time Radiance Field Rendering}. ACM Transactions on Graphics  \textbf{42}(4) (July 2023)

\bibitem{le2024voicebox}
Le, M., Vyas, A., Shi, B., Karrer, B., Sari, L., Moritz, R., Williamson, M., Manohar, V., Adi, Y., Mahadeokar, J., et~al.: {Voicebox: Text-Guided Multilingual Universal Speech Generation at Scale}. Advances in Neural Information Processing Systems  \textbf{36} (2024)

\bibitem{lewis2020retrieval}
Lewis, P., Perez, E., Piktus, A., Petroni, F., Karpukhin, V., Goyal, N., K{\"u}ttler, H., Lewis, M., Yih, W.t., Rockt{\"a}schel, T., et~al.: {Retrieval-Augmented Generation for Knowledge-Intensive NLP Tasks}. Advances in Neural Information Processing System  \textbf{33},  9459--9474 (2020)

\bibitem{li2024enhancing}
Li, J., Yuan, Y., Zhang, Z.: {Enhancing LLM Factual Accuracy with RAG to Counter Hallucinations: A Case Study on Domain-Specific Queries in Private Knowledge-Bases}. arXiv preprint \href{https://arxiv.org/abs/2403.10446}{arXiv:2403.10446}  (2024)

\bibitem{flame2017}
Li, T., Bolkart, T., Black, M.J., Li, H., Romero, J.: {Learning a Model of Facial Shape and Expression From 4D Scans}. ACM Trans. Graph.  \textbf{36}(6) (Nov 2017). \doi{10.1145/3130800.3130813}

\bibitem{li2024animatablegaussians}
Li, Z., Zheng, Z., Wang, L., Liu, Y.: {Animatable Gaussians: Learning Pose-dependent Gaussian Maps for High-fidelity Human Avatar Modeling}. In: Proceedings of the IEEE/CVF Conference on Computer Vision and Pattern Recognition (CVPR) (2024)

\bibitem{SMPL}
Loper, M., Mahmood, N., Romero, J., Pons-Moll, G., Black, M.J.: {SMPL: A Skinned Multi-Person Linear Model}. ACM Trans. Graph.  \textbf{34}(6) (Oct 2015). \doi{10.1145/2816795.2818013}

\bibitem{mildenhall2020nerfrepresentingscenesneural}
Mildenhall, B., Srinivasan, P.P., Tancik, M., Barron, J.T., Ramamoorthi, R., Ng, R.: {NeRF: Representing Scenes as Neural Radiance Fields for View Synthesis}. Commun. ACM  \textbf{65}(1),  99–106 (Dec 2021). \doi{10.1145/3503250}

\bibitem{TensorRT}
{Nvidia}: {TensorRT}. \url{https://github.com/NVIDIA/TensorRT}, accessed: 2025-03-20

\bibitem{pang2023dpedisentanglementposeexpression}
Pang, Y., Zhang, Y., Quan, W., Fan, Y., Cun, X., Shan, Y., Yan, D.M.: {DPE: Disentanglement of Pose and Expression for General Video Portrait Editing}. In: Proceedings of the IEEE/CVF Conference on Computer Vision and Pattern Recognition (CVPR). pp. 427--436 (June 2023)

\bibitem{SMPLX}
Pavlakos, G., Choutas, V., Ghorbani, N., Bolkart, T., Osman, A.A.A., Tzionas, D., Black, M.J.: {Expressive Body Capture: 3D Hands, Face, and Body from a Single Image}. In: Proceedings IEEE Conf. on Computer Vision and Pattern Recognition (CVPR) (2019)

\bibitem{wakeword}
{Picovoice}: {Porcupine}. \url{https://github.com/Picovoice/porcupine}, accessed: 2025-03-26

\bibitem{qwen2.5}
{Qwen Team}: {Qwen2.5: A Party of Foundation Models}. \url{https://qwenlm.github.io/blog/qwen2.5/} (September 2024)

\bibitem{radford2023whisper}
Radford, A., Kim, J.W., Xu, T., Brockman, G., McLeavey, C., Sutskever, I.: {Robust Speech Recognition via Large-Scale Weak Supervision}. In: International Conference on Machine Learning. pp. 28492--28518. PMLR (2023)

\bibitem{reimers-2019-sentence-bert}
Reimers, N., Gurevych, I.: {Sentence-BERT: Sentence Embeddings using Siamese BERT-Networks}. In: Proceedings of the 2019 Conference on Empirical Methods in Natural Language Processing. Association for Computational Linguistics (11 2019)

\bibitem{gptsovits}
{RVC-Boss}: {GPT-SoVITS}. \url{https://github.com/RVC-Boss/GPT-SoVITS}, accessed: 2025-03-20

\bibitem{uncannyvalley}
Seyama, J., Nagayama, R.S.: {The Uncanny Valley: Effect of Realism on the Impression of Artificial Human Faces}. Presence  \textbf{16}(4),  337--351 (2007). \doi{10.1162/pres.16.4.337}

\bibitem{shao2025degas}
Shao, Z., Wang, D., Tian, Q.Y., Yang, Y.D., Meng, H., Cai, Z., Dong, B., Zhang, Y., Zhang, K., Wang, Z.: {DEGAS: Detailed Expressions on Full-Body Gaussian Avatars}. In: Proceedings of the International Conference on 3D Vision (3DV) (2025)

\bibitem{shao2024splattingavatarrealisticrealtimehuman}
Shao, Z., Wang, Z., Li, Z., Wang, D., Lin, X., Zhang, Y., Fan, M., Wang, Z.: {SplattingAvatar: Realistic Real-Time Human Avatars with Mesh-Embedded Gaussian Splatting}. In: Proceedings of the IEEE/CVF Conference on Computer Vision and Pattern Recognition (CVPR) (2024)

\bibitem{shen2024naturalspeech2}
Shen, K., Ju, Z., Tan, X., Liu, E., Leng, Y., He, L., Qin, T., sheng zhao, Bian, J.: {NaturalSpeech 2: Latent Diffusion Models are Natural and Zero-Shot Speech and Singing Synthesizers}. In: International Conference on Learning Representations. pp. 1--25 (2024)

\bibitem{Imitator}
Thambiraja, B., Habibie, I., Aliakbarian, S., Cosker, D., Theobalt, C., Thies, J.: {Imitator: Personalized Speech-driven 3D Facial Animation}. In: Proceedings of the IEEE/CVF International Conference on Computer Vision (ICCV). pp. 20621--20631 (October 2023)

\bibitem{wang2023survey}
Wang, C., Liu, X., Yue, Y., Tang, X., Zhang, T., Jiayang, C., Yao, Y., Gao, W., Hu, X., Qi, Z., et~al.: {Survey on Factuality in Large Language Models: Knowledge, Retrieval and Domain-Specificity}. arXiv preprint \href{https://arxiv.org/abs/2310.07521}{arXiv:2310.07521}  (2023)

\bibitem{wang2022faceversefinegraineddetailcontrollable3d}
Wang, L., Chen, Z., Yu, T., Ma, C., Li, L., Liu, Y.: {FaceVerse: a Fine-grained and Detail-controllable 3D Face Morphable Model from a Hybrid Dataset}. In: IEEE Conference on Computer Vision and Pattern Recognition (CVPR) (June 2022)

\bibitem{wang2024domainrag}
Wang, S., Liu, J., Song, S., Cheng, J., Fu, Y., Guo, P., Fang, K., Zhu, Y., Dou, Z.: {DomainRAG: A Chinese Benchmark for Evaluating Domain-specific Retrieval-Augmented Generation}. arXiv preprint \href{https://arxiv.org/abs/2406.05654}{arXiv:2406.05654}  (2024)

\bibitem{xie2024towards}
Xie, T., Rong, Y., Zhang, P., Liu, L.: {Towards Controllable Speech Synthesis in the Era of Large Language Models: A Survey}. arXiv preprint \href{https://arxiv.org/abs/2412.06602}{arXiv:2412.06602}  (2024)

\bibitem{yang2024instructtts}
Yang, D., Liu, S., Huang, R., Weng, C., Meng, H.: {InstructTTS: Modelling Expressive {TTS} in Discrete Latent Space With Natural Language Style Prompt}. IEEE/ACM Transactions on Audio, Speech, and Language Processing  \textbf{32},  2913--2925 (2024)

\bibitem{yang2024srag}
Yang, H., Zhang, M., Wei, D., Guo, J.: {SRAG: Speech Retrieval Augmented Generation for Spoken Language Understanding}. In: 2024 IEEE 2nd International Conference on Control, Electronics and Computer Technology (ICCECT). pp. 370--374. IEEE (2024)

\bibitem{yang2020facescapelargescalehighquality}
Yang, H., Zhu, H., Wang, Y., Huang, M., Shen, Q., Yang, R., Cao, X.: {FaceScape: A Large-Scale High Quality 3D Face Dataset and Detailed Riggable 3D Face Prediction}. In: IEEE/CVF Conference on Computer Vision and Pattern Recognition (CVPR) (June 2020)

\bibitem{zheng2023pointavatardeformablepointbasedhead}
Zheng, Y., Yifan, W., Wetzstein, G., Black, M.J., Hilliges, O.: {PointAvatar: Deformable Point-based Head Avatars from Videos}. In: Proceedings of the IEEE/CVF Conference on Computer Vision and Pattern Recognition (CVPR) (2023)

\end{thebibliography}

\end{document}